\def\year{2020}\relax
\documentclass[letterpaper]{article}
\usepackage{aaai20}
\usepackage{times}
\usepackage{helvet}
\usepackage{courier}
\usepackage[hyphens]{url}
\usepackage{graphicx}
\usepackage{graphicx}
\usepackage{amsmath}
\usepackage{amsfonts}

\newcommand{\E}{\mathbb{E}}
\DeclareMathOperator*{\argmax}{arg\,max}

\title{
    Scaling All-Goals Updates in Reinforcement Learning\\
    Using Convolutional Neural Networks
}

\author{
    Fabio Pardo, Vitaly Levdik, Petar Kormushev\\
    Robot Intelligence Lab, Imperial College London, United Kingdom\\
    \{f.pardo, v.levdik, p.kormushev\}@imperial.ac.uk
}

\begin{document}

\maketitle

\begin{abstract}
    Being able to reach any desired location in the environment can be a valuable asset for an agent. Learning a policy to navigate between all pairs of states individually is often not feasible. An \textit{all-goals updating} algorithm uses each transition to learn Q-values towards all goals simultaneously and off-policy. However the expensive numerous updates in parallel limited the approach to small tabular cases so far. To tackle this problem we propose to use convolutional network architectures to generate Q-values and updates for a large number of goals at once. We demonstrate the accuracy and generalization qualities of the proposed method on randomly generated mazes and Sokoban puzzles. In the case of on-screen goal coordinates the resulting mapping from frames to \textit{distance-maps} directly informs the agent about which places are reachable and in how many steps. As an example of application we show that replacing the random actions in $\varepsilon$-greedy exploration by several actions towards feasible goals generates better exploratory trajectories on Montezuma's Revenge and Super Mario All-Stars games.
\end{abstract}

\section{Introduction}

Reinforcement learning (RL) \cite{sutton1998book} environments can typically be separated into goal-reaching and reward-based types. In the former case, tasks like navigating a maze or moving an object to a target location directly justify the use of agents that are conditioned on both the observations of the world and the goals. In the latter case of RL environments, maximizing the discounted sum of future rewards provides a more subtle objective that can describe any given task. While learning can be more challenging in this case, it can sometimes be greatly simplified by learning a task-agnostic goal-directed policy in combination with a higher-level task-dependent one. \cite{bakker2004hierarchical,kulkarni2016hierarchical,vezhnevets2017feudal,peng2017deeploco}.

In both cases, learning to reach each goal independently would be very inefficient. However, because the goals conditioning the policy do not modify the dynamics of the environment, off-policy algorithms can greatly improve the sample efficiency by using \textit{goal relabeling}. When replaying a past transition, this technique consists of substituting the given goal with another valid goal, in order to evaluate the action with respect to the new goal. Multiple strategies exist for selecting these goals. For example, within the list of future states visited in the episode from which the transition originated, as was proposed in Hindsight Experience Replay (HER) \cite{andrychowicz2017hindsight}, randomly \cite{schaul2015universal,veeriah2018many} or from a generative model \cite{nair2018visual}. However, in the case where all the possible goals can be enumerated, a single transition can be maximally used by updating the policy towards \textit{all-goals}.

A traditional goal-conditioned network requires one forward pass per goal which can be intractable in the case of large goal sets. Instead, we propose to use a network producing Q-values for all combinations of actions and goals with a single observation in input. Furthermore, we propose to use convolutional neural networks (ConvNets) to scale to a large number of outputs simultaneously, exploiting the correlations between neighbouring goals.

We evaluate the accuracy and generalization properties of the proposed network on tasks consisting in finding the shortest path towards all points in random mazes and solving random Sokoban puzzles. Our results show that the proposed approach converges faster and achieves better final performance while being more stable than the the goal-in-input approach. We then demonstrate that the proposed network learns well in more visually complex and difficult to control environments. On Montezuma's Revenge we show that the capacity to query this large number of Q-values can allow an agent to explore well by selecting random but feasible goals. Finally, on Super Mario All-Stars, we show that when coupled with a task-learner DQN agent \cite{mnih2015human} the exploration method significantly improves the performance.

The source code and videos are available on the website: \url{https://sites.google.com/view/q-map-rl}.

\section{Related work}

\subsection{Goal relabeling strategies}

Goal relabeling strategies rely on the fact that a single transition can be used to update the policy towards any goal. A few different methods for selecting these goals have been proposed, some of which are explained bellow.

\paragraph{Future states}

In Hindsight Experience Replay (HER) \cite{andrychowicz2017hindsight}, the goals are sampled from the list of subsequent states traversed in the remainder of the episode when selecting a transition. This allows to quickly propagate successful examples of goal reaching as all of the required transitions are present in the recorded episode. However, this approach limits the scope of goals considered.

\paragraph{Random states}

A simpler approach is to select goals uniformly over the set of goals which can be known or discovered through interaction \cite{schaul2015universal,veeriah2018many}. This approach allows to learn to reach any known goal if trajectories can be discovered in the set of transitions or the model has enough capacity to generalize. One downside however, is that many goals can be unreachable from a given state or not useful and the value function can easily bootstrap from incorrect values.

\paragraph{Imagined goals from a generative model}

If the set of possible states is unknown and we wish to get more diverse goals than the ones discovered so far, it is possible to learn a generator from which the goals can be sampled. For example, a variational autoencoder (VAE) \cite{kingma2014vae} can be trained to learn the distribution of states and used to generate new plausible goals \cite{nair2018visual}. This approach can lead to more general-purpose goal-reaching policies but relies on the accuracy of the learned generator. Therefore, a mixture of imagined and random goals can be more advantageous \cite{nair2018visual}.

\paragraph{All goals}

The previous approaches are sometimes referred to as \textit{many-goals learning} \cite{veeriah2018many} and are general enough to work with continuous and unknown goal spaces. However, many tasks of interest have a known finite number of goals. In which case, the best use of a single transition can be achieved by updating the policy towards all goals, hence the name all-goals updates. For example, an independent Q-value for each of the goals can be learned separately with tabular Q-learning \cite{kaelbling1993learning}. The main issue with this approach is that the number of updates scales linearly with the number of goals.

\subsection{Using ConvNets to represent value functions}

Training several function approximations in parallel quickly becomes impractical in the case of representing many value functions. Multiple works showed how a shared torso fully connected to multiple heads can be used to learn several value functions \cite{osband2016deep,van2017hybrid,serkan2018intentional,tavakoli2018action}. A few approaches aim at representing value functions using convolutional neural networks, some of which are described bellow.

\paragraph{RL with unsupervised auxiliary tasks}
In UNREAL \cite{jaderberg2017unreal}, the main policy is trained along multiple unsupervised auxiliary policy heads to help generate useful features. The pixel-control task trains the agent to predict which actions generate the most change in the input frames. The head specific to this task uses transposed convolutions to generate a three dimensional array of Q-values. These values were however never used to control the agent and were not specialised in goal-reaching.

\paragraph{Value Iteration Networks}
A VIN module \cite{tamar2016value} creates a reward frame which is fed into a convolutional layer generating Q-values which are then max-pooled along the action channel to produce a frame which is iteratively fed into the module until it represents the state-values of the optimal policy. An attention mechanism is then applied on the value frame to create features for a policy trained via reinforcement learning. A VIN module can in theory perform planning as demonstrated on goal-reaching tasks. While VIN is related to our model it has many differences. First, it does not directly use the generated values to act. Second, for goal-reaching tasks, those values are the expected returns when starting from each state location for a goal given in input while our approach represents the values of starting from a given state to reach all the possible goals. Finally, VIN performs planning by iteratively recomputing the values at every step while our approach is solely based on learning a direct mapping from inputs to values.

\section{Background}

\paragraph{Q-values}
We consider the standard reinforcement learning framework \cite{sutton1998book}, in which an agent interacts sequentially with its environment, formalized as a Markov Decision Process (MDP) with state space $\mathcal{S}$, action space $\mathcal{A}$, reward function $r(s,a,s')$ and state-transition function $p(s'|s,a)$. At each time step $t$, the agent generates an action $a_t$ sampled from its policy $\pi(a_t|s_t)$ conditioned on the state $s_t$. The environment responds by providing a new state $s_{t+1}$ and a reward $r_{t+1}$. Some states can be terminal, meaning that no more interaction is possible after reaching them, which can be simply considered as a deadlock state that only transitions to itself, providing no reward.

The action-value function of the policy is defined as:
\begin{equation*}
    Q^\pi(s,a) = \E_{s' \sim p, a' \sim \pi} \big[ r(s,a,s') + \gamma Q^\pi(s',a') \big]
\end{equation*}
It indicates the quality (Q-value) of each possible immediate action when following the policy afterwards.

\paragraph{Q-learning}
In the Q-learning algorithm \cite{watkins1992q}, the action-value function of the optimal policy $\pi^*$ is iteratively approximated by updating the estimated Q-values:
\begin{equation*}
    Q(s,a) \leftarrow (1-\alpha) Q(s,a) + \alpha \left(r + \gamma \max_{a'} Q(s',a') \right)
\end{equation*}
It uses previously experienced transitions $(s,a,s',r)$ and a learning rate $\alpha$. In $\varepsilon$-greedy exploration, this learned action-value function can be used to take greedy actions \mbox{$a = \argmax_a Q(s,a)$} or random actions uniformly \mbox{$a \sim \mathcal{U}(\mathcal{A})$} with probability $\varepsilon$. Finally, the fact that the target $r + \gamma \max_{a'} Q(s',a')$ does not rely on the policy used to generate the data, allows Q-learning to learn off-policy, efficiently re-using previous transitions stored in a replay buffer or generated by another mechanism.

\paragraph{Generalized and Universal Value Functions}
While an action-value function is usually specific to the rewards defining the task, the Generalized Value Functions (GVFs) \cite{sutton2011horde} $Q^\pi_g(s,a)$ are trained with pseudo-reward functions $r_g(s,a,s')$ that are specific to each goal $g$. The Horde architecture combines a large number of independent GVFs trained to predict the effect of actions on sensor measurements and can be simultaneously trained off-policy. The Universal Value Function Approximators (UVFAs) \cite{schaul2015universal} extend the concept of GVFs by adopting a unique action-value function $Q^\pi(s,a,g)$ parameterized by goals and states together, enabling interpolation and extrapolation between goals.

\section{Proposed model}

\subsection{Training and usage}

While UVFA-like architectures take observations and goals in input and generate Q-values for every action in output, our model, named Q-map, only takes observations in input and generates Q-values for every goal and action in output. For example, in the case of a grid of 2D goal coordinates, when a stack of observation frames are provided in input, another stack of 2D \textit{Q-frames} are generated where the rows and columns represent the goal locations and the number of frames represents the number of actions. Note that the height and width of the observations and generated frames can be different for example if the observations require more resolution as shown in some of the following experiments.

These output values represent Q-values in the context of the goal-reaching reward function awarding $1$ with episode-termination at the goal and $0$ otherwise. The discount factor $\gamma$ creates exponentially decaying values $\gamma^{k-1}$ indicating the number of steps $k$ to the goal. A value of $0$ indicates that the goal cannot be reached (for example in an obstacle) and $1$ means that the goal will be reached at the next time step.

For a given transition, only the output frame corresponding to the taken action is updated using a mean-squared error with a target frame. This target is generated as follows: First, a forward pass in the model is performed with the next observation in input. Then, the generated frames are maximized over the action dimension to generate a single frame representing a measure of the minimum expected number of steps towards each of the goals. This frame is then clipped to the range $(0, 1)$ to reduce the under- and over-estimations. Finally, the frame is discounted by $\gamma$ and the location reached at the next step is set to $1$. This procedure effectively uses one environmental transition to virtually create a set of independent one-step episodes, one per goal, with termination on success and partial-episode bootstrapping otherwise \cite{pardo2018time}.

To use the generated Q-frames to reach a specific goal, one only needs to take the vector of Q-values at the goal location and select the action of maximum value.

\begin{figure}
    \centerline{\includegraphics[width=\columnwidth]{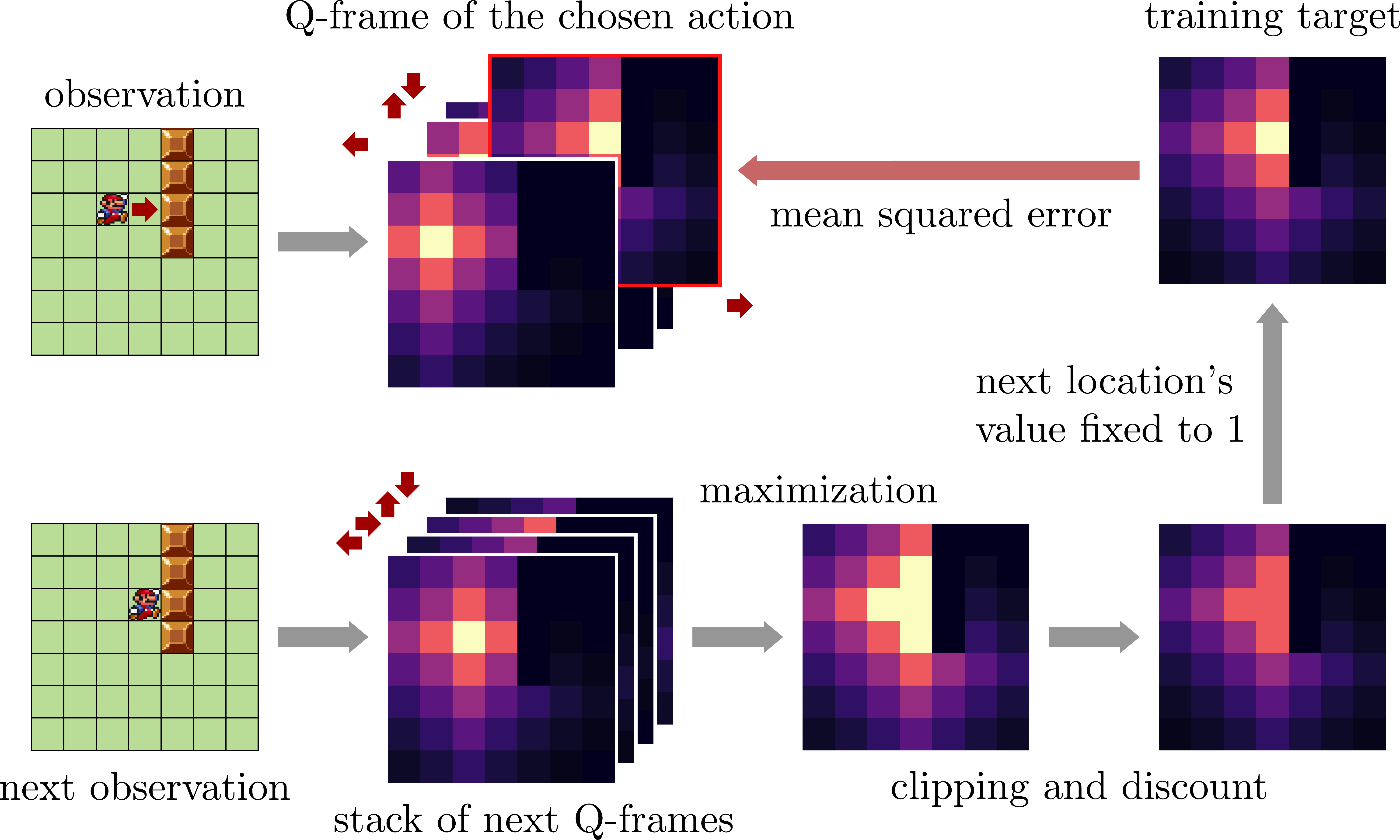}}
    \caption{Training process for the proposed model, updating the prediction towards all goals at once.}
    \label{fig:qmap_update}
\end{figure}

\subsection{Neural network architectures}

We considered multiple neural network architectures to support the proposed model. The first choice, which will be our baseline in the rest of the paper, is a UVFA-like network which takes a stack of observation and goal frames and generates the corresponding vector of Q-values, one for each action. To create a many-goals or all-goals update, the batch needs to consists of frames where the observation remains the same but the goal, represented by a one-hot frame, changes to cover all desired values. The batch output is then a set of vectors which can be reshaped to create the expected stack of Q-frames.

The second network uses an autoencoder-inspired architecture, with convolutions followed by fully-connected layers in turn followed by transposed convolutions. This network takes observation frames in input and generates the full Q-frames in output. The convolutional nature of this approach benefits from a potentially better capacity to share features and use correlations between the vision of the environment and the expected number of steps.

Finally, the last network architecture that we considered is composed solely of convolutions without compression and decompression. For example if the height and width of the Q-frames corresponds to the ones of the observations, strides $1$ and padding ``same'' are used to keep the shape of the feature maps constant, not loosing any localization information.

\subsection{Exploration with random goals}

As a simple application case for a Q-map we propose an exploration method that replaces the noisy random actions frequently used to explore in reinforcement learning with a sequence of steps towards a goal. Using the Q-frames produced via Q-map, the goal is selected within a estimated close proximity to the current position and the actions are chosen greedily in order to reach it. While we do not expect this exploration method to outperform specialised ones that use a variety of intrinsic signals, such as information gain \cite{kearns1999efficient,brafman2002r}, state visitation counts \cite{bellemare2016unifying,tang2017exploration} or prediction error \cite{stadie2015incentivizing,pathak2017curiosity}, it can be incorporated into most of those or used as a drop-in replacement for $\varepsilon$-greedy. It is worth noting that other works proposed to base the exploration on goal selection \cite{baranes2013active,florensa2017reverse,pere2018unsupervised,colas2018gep} but to the best of our knowledge none of them relied on generating large and consistent steps in the environment using an auxiliary goal-reaching policy.

\section{Experiments}
\label{experiments}

In sections \ref{mazes} and \ref{sokoban} we aim to evaluate the accuracy, training time and generalization properties of the proposed Q-map model on gridworld environments while in sections \ref{montezuma} and \ref{mario} we test Q-map in more visually complex environments and propose an application to exploration in reinforcement learning. In all of the experiments we use $\gamma = 0.9$ for the goal-reaching Q-functions and the neural networks are described using the notations: conv(filters, kernel sizes, strides) for convolutions (with padding ``same'' unless stated otherwise), deconv2d for transposed convolutions and dense(units) for dense layers. Elu activation functions are used for every layer except for the output ones. For more details, the full source code of the experiments is available at \url{https://sites.google.com/view/q-map-rl}.

\subsection{Pathfinding in random mazes}
\label{mazes}

\begin{figure}[t!]
    \centerline{\includegraphics[width=\columnwidth]{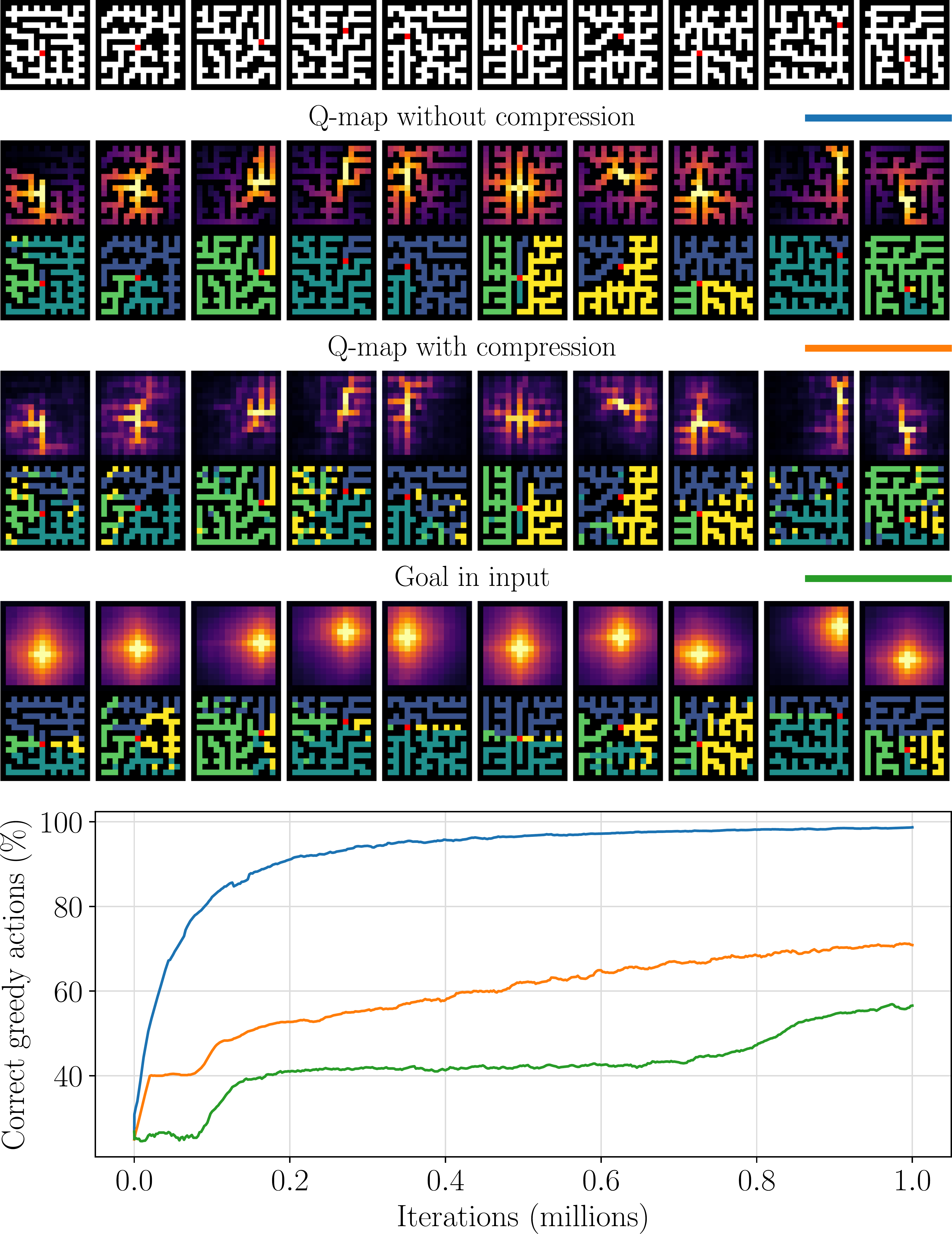}}
    \caption{Learning to solve random mazes. Top: $10$ unseen random mazes, then for each agent: the max Q-frames (maximized over the action dimension) and the first greedy action towards each goal. Bottom: percentage of correct greedy actions to take towards each possible goal. The Q-map architecture performs better than the baseline, in particular with the no-compression network.}
    \label{fig:mazeworld_results}
\end{figure}

The environment consists of a single pixel that can be moved in cardinal directions in a $16 \times 16$ area with traversable pathways surrounded by walls and generated such that any two points in the maze are only connected by one path. Actions towards walls result in the pixel remaining stationary. The observations consist of a stack of $16 \times 16$ RGB frames of the full view of the maze. The background is white, walls are black while the controlled pixel is in red. For the baseline model we use a green pixel to represent the goal.

We consider three dueling double DQN architectures for evaluation \cite{wang2016dueling,hasselt2010double,mnih2015human}. The baseline architecture ``goal-in-input'' uses 2$\times$conv(64, 4, 2)-dense(512) layers followed by a dense(256)-dense(4) advantage branch and a dense(256)-dense(1) state-value branch. The online network (not counting the target one) uses approximately $860$K parameters. The second architecture Q-map ``with compression'' uses 2$\times$conv(64, 4, 2)-dense(512)-dense(1024) layers followed by a deconv(64, 4, 2)-deconv(4, 4, 2)  advantage branch and a deconv(64, 4, 2)-deconv(1, 4, 2) state-value branch. The online network uses approximately $1,260$K parameters. The third architecture Q-map ``without compression'' is composed of 4$\times$conv(64, 4, 1) layers followed by a 3$\times$deconv(64, 4, 1)-deconv(4, 4, 1) advantage branch and 3$\times$deconv(64, 4, 1)-deconv(1, 4, 1) state-value branch. The online network uses approximately $800$K parameters.

Instead of letting the agents interact with the environment we choose to generate a training and testing sets that are identical between the agents in order to remove any exploration bias when comparing the quality of the Q-frames generated. The training set is comprised of all possible transitions in $2,317$ different mazes for a total of $1,000,220$ transitions. The testing set is comprised of all possible starting points in $10$ new mazes for a total of $1,075$ observations. The agents are trained with batches of $50$ random transitions from the training set. The evaluation metric is the ``success rate'' which is a proportion of state-goal pairs where the agents correctly predict the greedy action towards all feasible goals, obtained by using the $\argmax$ operator over the action dimension.

The Figure \ref{fig:mazeworld_results} shows the training success rate curves for the architectures as well as examples of the learned Q-frames for the test mazes. In the solid-color filled maze images the color of a pixel depicts the action-choice made by the corresponding agent towards that pixel from the given location. Logically all pathways towards a single direction from the agent's location should have the same color. The Q-map ``without compression'' architecture clearly outperforms the alternative, achieving $99\%$ success rate at $1M$ training iterations. The ``Goal in input'' architecture has learned what one could consider a naive first-pass solution to the task - that is a fading gradient centered on the agent's location in the maze, but failed to recognize walls of the maze. This is a common phase in training of all architectures, and indicates that the approach could require many more training iterations to improve further. The Q-map ``with compression'' based architecture performed significantly better but still struggled to produce sufficiently sharp Q-frames, unable to determine correct action choices outside of close proximity to the agent. It is worth noting that the Q-values for the maze walls are not specifically masked and the Q-map successfully learns to decay them to $0$. Videos showing the Q-frames and actions through the training of all three architectures are available on the website.

\subsection{Solving Sokoban puzzles}
\label{sokoban}

\begin{figure}
    \centerline{\includegraphics[width=\columnwidth]{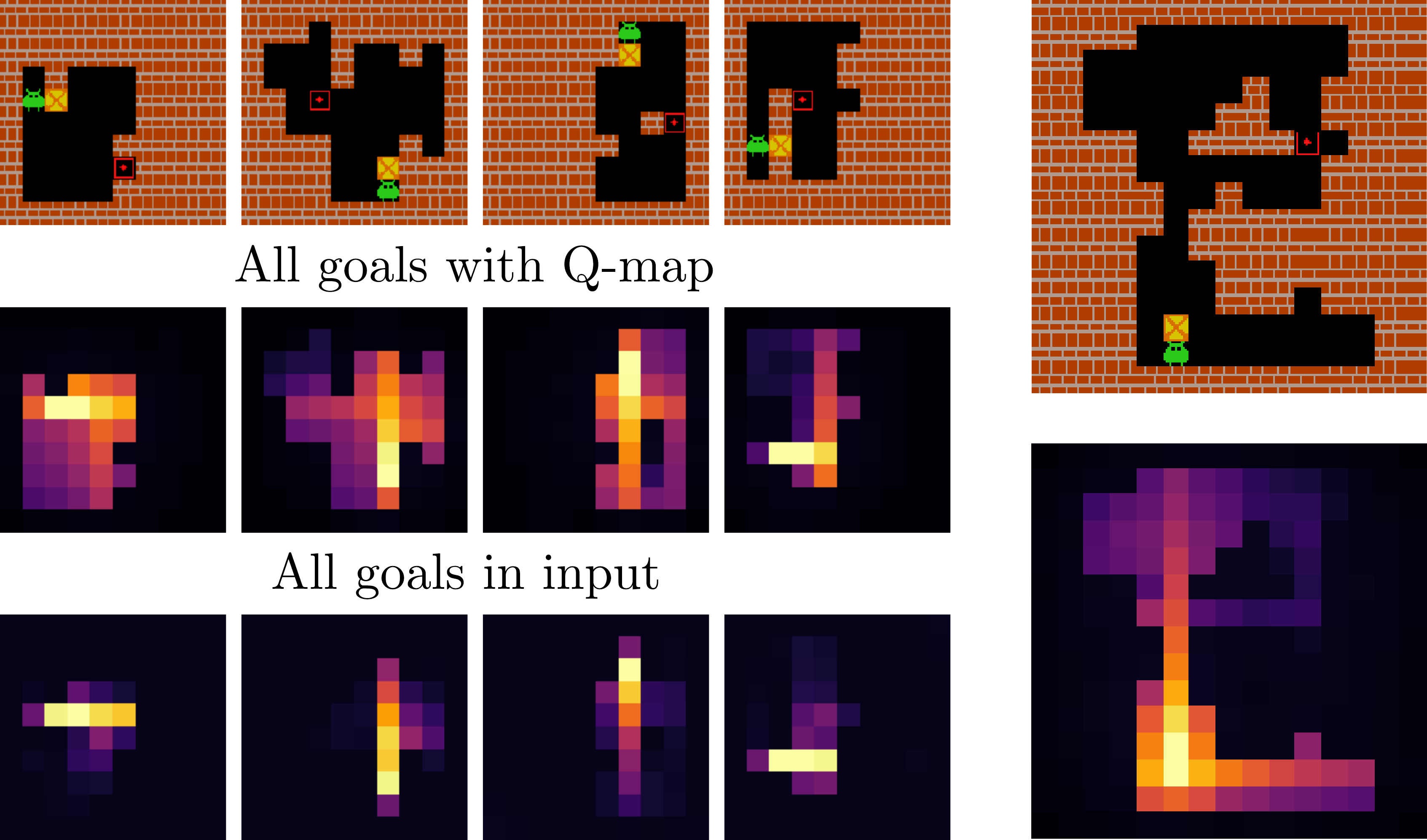}}
    \caption{Left: Comparison between the max Q-frames learned by Q-map and the all-goals in input approaches on unseen $10 \times 10$ puzzles. Right: A separate example from a Q-map trained on $15\times15$ levels.}
    \label{fig:sokoban_qframes}
\end{figure}

In this section we consider a notoriously difficult environment suited for planning \cite{racaniere2017imagination}: Sokoban, from Gym-Sokoban \cite{schrader2018sokoban}. A significant difference to the previously considered Maze environment is that the goal is specified for a box which has to be carefully pushed by an agent-controlled avatar.

Similarly to the previous experiments, we compare the Q-map ``without compression'' model with two goal-in-input baselines. The two baselines differ by their goal-relabeling strategies: one uses random goals while the other one uses all goals. All the models are trained on the same batch size of $100$. For the all-goals updates model, because there are $100$ possible goals, a batch contains a single observation replicated $100$ times with each of the possible goals.

Both of the baseline models uses a 5$\times$conv(64, 5, 1)-2$\times$conv(64, 5, 1, valid)-dense(256)-dense(4) network, while the Q-map ``with compression'' uses a 7$\times$conv(64, 5, 1)-conv(4, 5, 1) network. We keep the total number of parameters roughly the same between the models with approximately $688$K for the baselines and $626$K for the Q-map.

\begin{figure}
    \centerline{\includegraphics[width=\columnwidth]{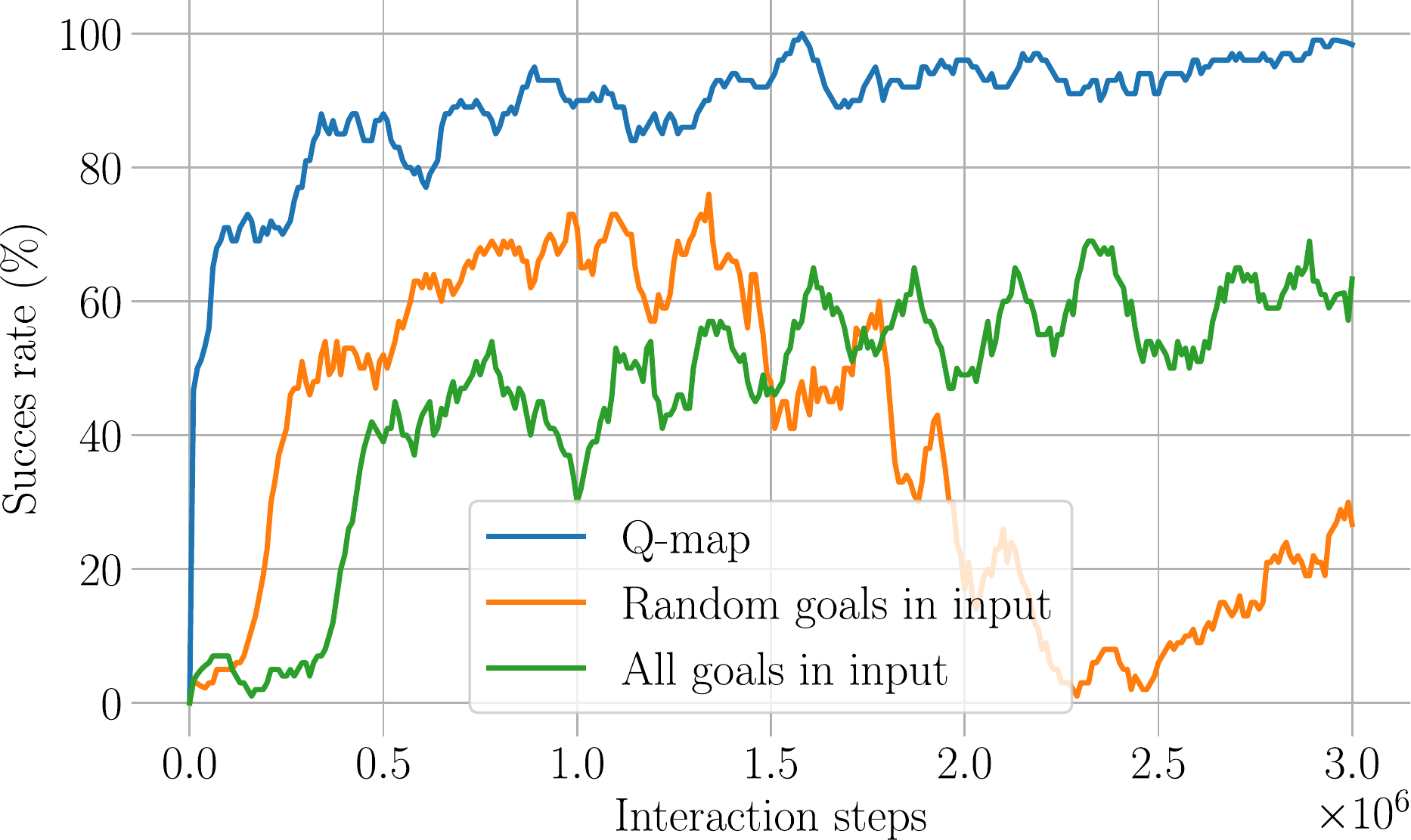}}
    \caption{Success rate on unseen Sokoban puzzles. Q-map performs better than the goal-in-input baselines. The random-goals baseline is particularly unstable.}
    \label{fig:sokoban_score}
\end{figure}

We generate a unique $10 \times 10$ Sokoban level per episode including for evaluation. The models are provided with $10 \times 10$ RGB frames representing the walls, box and the agent's locations in red, green and blue respectively. In contrast to the Maze environment, we generate data by performing exclusively random actions and storing the state transitions in an unlimited buffer. This results in a less uniform training set that is more representative of an agent interacting with the environment, while using the same seeds generates identical transitions between agents.

Multiple generated Q-frames examples are given in Figure \ref{fig:sokoban_qframes}. The Q-map model was able to learn better representations showing how the box should be moved even in complex scenarios requiring a sequence of actions with long-term dependencies. Figure \ref{fig:sokoban_score} shows the success rate of each of the considered models measured as the proportion of goals reached with greedy actions. Each point on the graph represents an average of $10$ individual evaluation runs. The Q-map model straight away outperforms both of the baselines and keeps improving until almost perfect performance in a very stable way. Of the two baselines, the one trained with random-goals initially seems to learn faster but then deteriorates, suggesting that learning with all goals simultaneously is more stable.

Finally, we note that the Q-map model could in theory support multiple boxes by increasing the dimensionality of the outputs to account for the combinatorial nature of the problem. We however leave this experiment for future work.

\begin{figure}
    \includegraphics[width=\columnwidth]{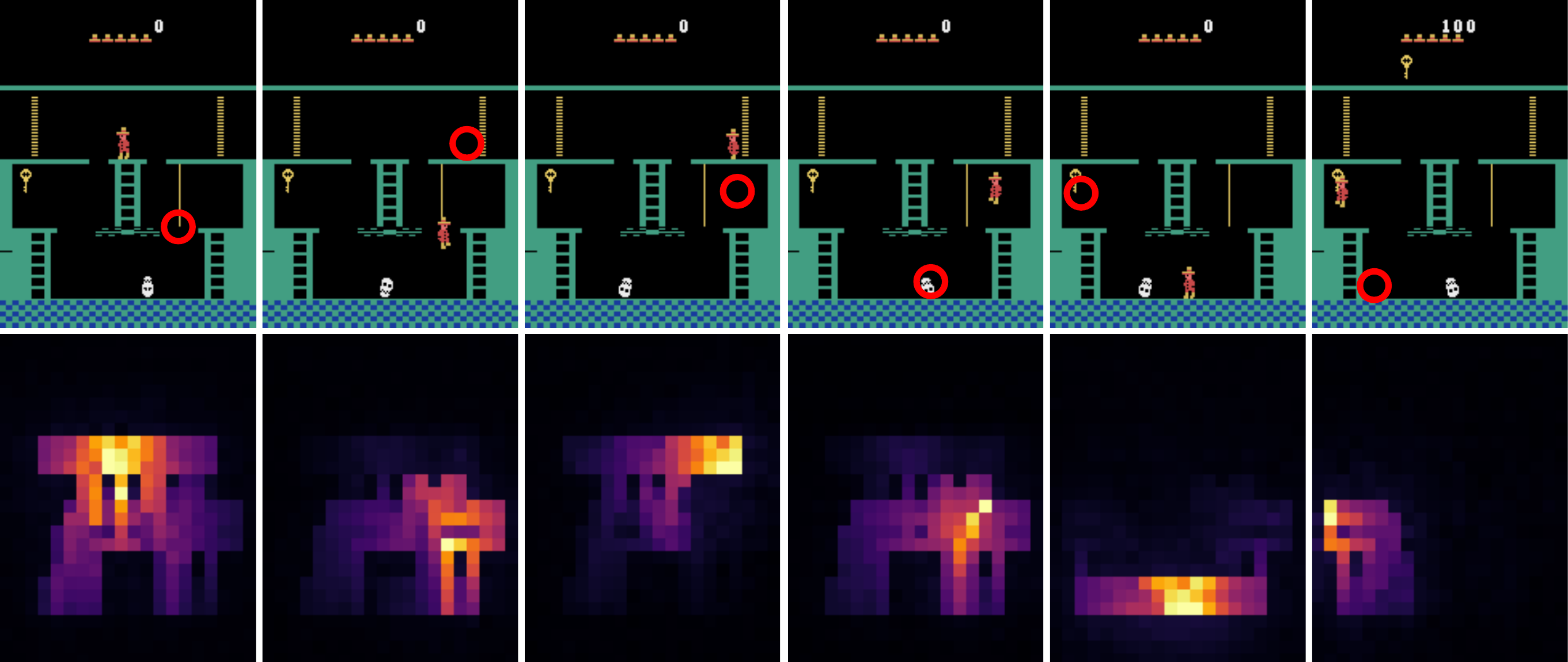}
    \caption{A random sequence of goal-reaching steps successfully exploring most of the first room in the Montezuma's Revenge game. The goals are shown with circles and the corresponding learned max Q-frames accurately represent reachable places.}
    \label{fig:montezuma_exploration}
\end{figure}

\subsection{Montezuma's Revenge}
\label{montezuma}

So far we have demonstrated the efficiency of the proposed Q-map model in a selection of gridworld problems which, while challenging to solve, have very simple state spaces and action dynamics. In this section we use the Montezuma's Revenge (Atari 2600) game to evaluate the learning capability of Q-map and roughly compare the exploration boundaries between a random-action policy and a random-goal policy that utilizes Q-map for action selection. Environmental rewards are ignored and no task-learner agent is present in this experiment.

The actions are limited to ``no-op, left, right, jump, jump-left, jump-right, down'' and are repeated four times. The game frames are zero-padded to reach a resolution of $160 \times 224$ before being scaled with a factor of $1/4$. Three grey-scale $40 \times 56$ frames, spaced by two steps are stacked to produce the observations while another scaling factor of $1/2$ is used for the coordinates, limiting the Q-frames to $20 \times 28$.

We use a Q-map ``with compression'' to accommodate the resolution discrepancy between the observations and the output. The network uses conv(32,8,2)-conv(32,6,2)-conv(64,4,2)-dense(1024)-dense(1024) layers followed by a deconv(32, 4, 2)-deconv(32, 6, 2)-deconv(4, 8, 1) advantage branch and a deconv(32, 4, 2)-deconv(32, 6, 2)-deconv(1, 8, 1) state-value branch. A random goal is chosen within $15$ to $30$ predicted steps from the agent's current position. An individual goal-directed trajectory terminates upon either reaching the goal or exceeding $150\%$ of the original predicted number of steps. Furthermore, there is a chance to take a random action decayed linearly from $0.1$ to $0.05$.

Qualitatively, the learned Q-frames are sufficiently accurate for the random-goal policy to be able to navigate much further through the environment and even actively avoid the contact with the skull on the way towards a goal. Example Q-frames are shown in Figure \ref{fig:montezuma_exploration}. We also tracked the number of keys picked up by both of the policies during the training, with the random policy only reaching the key once in the $5$ million steps, while the random-goal policy first reached the key at the $1.2$ million steps and overall $398$ times.

\subsection{Super Mario All-Stars}
\label{mario}

Finally, the proposed exploration method is used as a drop-in replacement for the $\varepsilon$-greedy exploration in a DQN agent on the Super Mario All-Stars game (SNES) \cite{openai2018retro}. The actions are limited to ``no-op, left, right, up-left, up, up-right'' and are repeated four times. The rewards from the game are divided by $100$ with $0$ bonus for moving to the right or penalty for game overs. Terminations by touching enemies or falling into pits and the coordinates of Mario and of the scrolling window are extracted from the RAM. Episodes are naturally limited by the timer of $400$ seconds present in the game, which corresponds to $2,402$ steps. Observations consist in three $56 \times 64$ grayscale frames spaced by two steps and the goal space is scaled down to $32 \times 28$.

\begin{figure}
    \includegraphics[width=\linewidth]{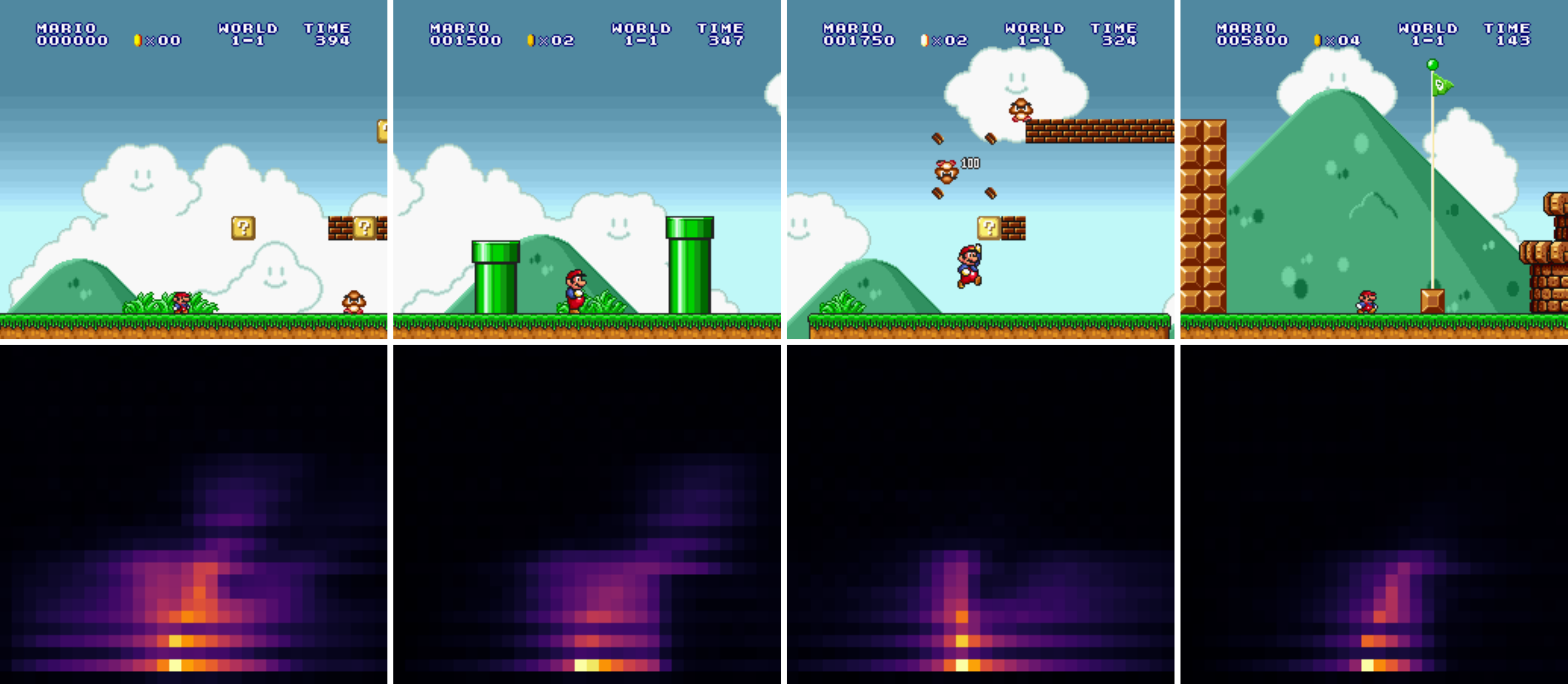}
    \caption{Example of learned max Q-frames on Super Mario All-Stars. The Q-map successfully take into account the obstacles. The horizontal patterns are due to the action repeat.}
    \label{fig:mario_qframes}
\end{figure}

We used the network described in Section \ref{montezuma} for Q-map. The network used for DQN has conv(32,8,2)-conv(32,6,2)-conv(64,4,2) layers followed by a dense(1024)-dense(6) advantage branch and a dense(1024)-dense(1) state-value branch. Furthermore, to account for the movement of the screen in the game, the target Q-frames are shifted accordingly before performing the updates.

\begin{figure*}
    \includegraphics[width=\linewidth]{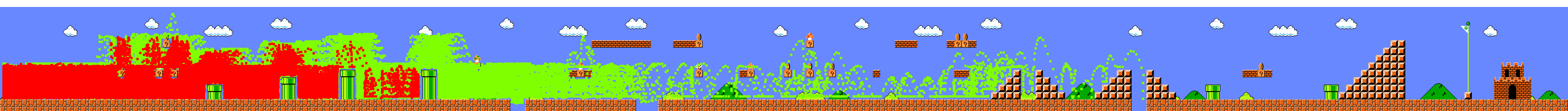}
    \includegraphics[width=\linewidth]{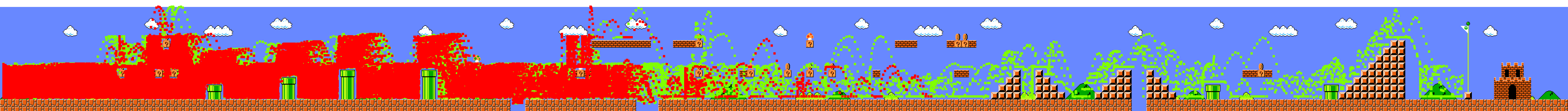}
    \caption{Coordinates visited after $2$ million steps on the first level of Super Mario All-Stars. First image: Random walk (red) compared against the proposed Q-map random-goal walk (green). Second image: DQN using $\varepsilon$-greedy (red) compared against DQN with the proposed exploration (green). In both cases, Q-map allows to explore significantly further}
    \label{fig:mario_exploration}
\end{figure*}

Similarly to Section \ref{montezuma}, we initially compare a random-goal Q-map based policy with a random-action policy. The states visited during $2$ million steps of interaction are displayed in Figure \ref{fig:mario_exploration}). As can be seen, the proposed exploration method covers significantly larger area, allowing the agent to almost reach the end of the level without using any environmental or intrinsic reward.

\begin{figure}
    \centerline{\includegraphics[width=\columnwidth]{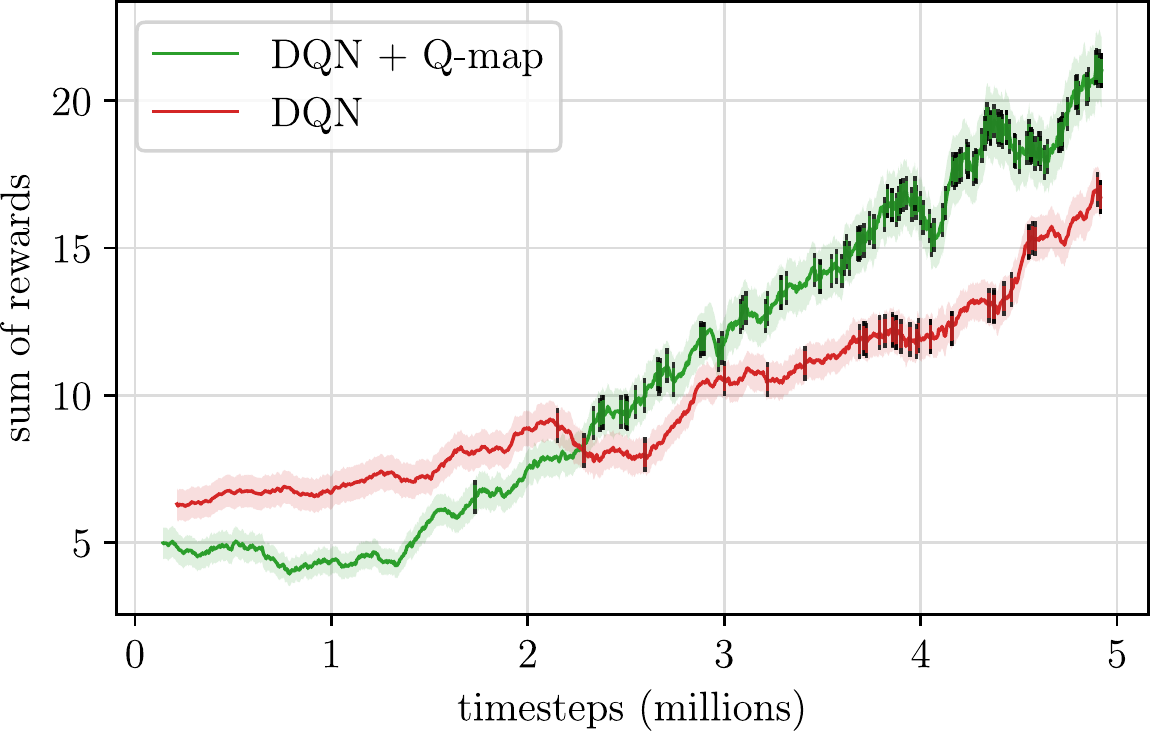}}
    \caption{Performance comparison between $\varepsilon$-greedy exploration (red), and the proposed exploration (green) with confidence intervals of $99$\%. The vertical bars indicate flags reached. The proposed agent significantly outperforms the baseline, reaching the flag earlier and more frequently.}
    \label{fig:mario_score}
\end{figure}

We then evaluate an augmented agent that is comprised of a Q-map-based exploration and a DQN \cite{mnih2015human} task-learner agent. For both this agent and the baseline we use an exploratory schedule that linearly decreases from $100\%$ to $5\%$. For the proposed agent, the random action probability is decreased from $10\%$ to $5\%$ over the course of the training. In order for the total proportion of exploration steps to match the planned schedule, the chance to start a new goal-reaching trajectory is then dynamically adjusted. Furthermore, to focus the exploration towards the task-learner policy a $50\%$ chance to select the goals with a first greedy action identical to the one from the task-learner is introduced.

To measure the performance of the proposed combined agent in terms of the sum of rewards collected per episode and the number of flags reached, we ran both agents for $5$ million steps with the same four seeds ($0$, $1$, $2$, $3$) and reported the results in Figure \ref{fig:mario_score}. The initial performance of the augmented agent is worse than the baseline, likely due to the fact that early rewards can easily be collected with random movements. In turn the longer exploration horizon of the augmented agent enables it to learn to progress through the level faster and learn to reach the final flag consistently. In total, the baseline reached the flag $9$ times while the proposed agent reached it $33$ times with a final performance $30\%$ higher. Finally, we also tested the capacity of the Q-map model to adapt to a different level and found that the learning was significantly faster when transferring a pre-trained Q-map model from one level to another than when training from scratch as visible in the videos.

\section{Discussion}

In the experiments using Montezuma's Revenge and Super Mario All-Stars we assumed that the location of the avatar was available to create the updates. While this can appear as a strong assumption, it is not unreasonable to imagine that an agent could learn to localize its avatar or other controllable objects \cite{moniz2019compression,sawada2018disentangling} and use this knowledge when training a Q-map model.

Furthermore, it is worth noting that the proposed approach could be extended beyond images in observations, such as angles and velocities of objects or point clouds and the Q-frames could be replaced by Q-tensors to represent larger coordinate spaces architecturally achieved by using multi-dimensional convolutions. This could, for example, enable an agent to control robotic arms or flying drones.

Beyond the examples demonstrated in Section \ref{experiments}, learning to reach coordinates with Q-map can be useful in many scenarios. For example in hierarchical reinforcement learning, a high-level agent could be rewarded to provide useful goals to a low-level Q-map model. Also, a count-based exploration method could be combined with the estimated distance from a Q-map to select close and less visited coordinates. Furthermore, learning Q-map in itself could be a signal for a curiosity-like exploration algorithm.

\section{Conclusion}

We proposed an all-goals Q-learning model that is computationally efficient and able to generalise to previously unseen goals. A single environmental transition is used to generate a value estimate for the full set of possible goals in a single forward pass in the network. This is achieved by utilising a convolutional architecture with the intuition that such a model would take advantage of the correlations between the input features and the similarities of the neighbouring goals in the environment. We have demonstrated that this approach significantly outperforms goal-in-input baselines and achieves nearly perfect success rate in gridworld maze pathfinding and Sokoban tasks. In addition we have shown that the Q-map model is capable to learn to navigate in visually complex environments such as Montezuma's Revenge and Super Mario All-Stars games. Finally we provide an example of an application where by replacing random $\varepsilon$-greedy exploration actions with random goal-directed trajectories we improve the performance of a DQN agent.

\section*{Acknowledgments}
The research presented in this paper has been supported by Dyson Technology Ltd. and computation resources were provided by Microsoft Azure.

\fontsize{9.4pt}{10.6pt}
\selectfont
\bibliographystyle{aaai}
\bibliography{main}

\end{document}